# Typologie des comportements utilisateurs : étude exploratoire des sessions de recherche complexe sur le Web

Claire Ibarboure, Ludovic Tanguy and Franck Amadieu

*Laboratoire CLLE - CNRS & Université de Toulouse Jean Jaurès*

**Abstract**
Dans le cadre de cette étude, nous proposons une approche exploratoire afin de proposer une typologie des comportements utilisateurs lors d'une session de recherche sur le Web. Nous décrivons une typologie fine à l'échelle de la session, reposant sur des variables génériques de la RI (e.g. nombre de requêtes) mais également sur l'étude de propositions thématiques (i.e. proposition avec un contenu sémantique distinct définie à partir de l'énoncé de recherche). Pour cela, nous travaillons à partir de données expérimentales nous permettant l'étude de la variation pour une même tâche donnée à différents utilisateurs (N=70). À la suite d'une analyse multidimensionnelle, nous proposons une typologie de 5 classes reposant sur le comportement des individus pendant le traitement d'une tâche de recherche complexe.

## 1. Introduction

Lorsque des utilisateurs sont confrontés à une même tâche complexe de Recherche d'Information (RI) sur le Web (i.e. une tâche nécessitant la recherche de plusieurs objectifs et par conséquent plusieurs requêtes pour répondre au besoin d'information), nous savons qu'ils ne réalisent pas une session de recherche identique. En effet, les utilisateurs sont souvent amenés à planifier leur recherche en définissant l'objectif et parfois les sous-objectifs nécessaires pour répondre au besoin de la tâche demandée [1]. Cependant, ces sous-objectifs ne sont pas traités de la même façon par tous les utilisateurs.

Nous avons été confrontés à cette variation lors de l'étude des comportements de 70 utilisateurs à travers l'analyse de leur session de recherche pour une même tâche. Nous nous sommes donc demandés s'il était possible de regrouper ces variations afin de proposer une typologie comportementale des utilisateurs lors du traitement d'une session de recherche. Cet objectif a soulevé plusieurs questions, notamment le choix des variables pour construire notre typologie.

Traditionnellement pour décrire une session de recherche, nous retrouvons un même noyau de variables : le nombre de requêtes formulées par l'utilisateur, le nombre de pages visitées ou encore la durée de la session, qui sont facilement observables et pouvant être extraites automatiquement. Cependant, ces variables-ci ne permettent pas de qualifier finement une

---

claire.ibarboure@univ-tlse2.fr (C. Ibarboure); ludovic.tanguy@univ-tlse2.fr (L. Tanguy); franck.amadieu@univ-tlse2.fr (F. Amadieu)

session de recherche. Autrement dit, avec ces variables, nous ne prenons pas en compte l'enchaînement des actions (e.g. les requêtes) et nous ne pouvons pas dégager une analyse globale de la session de recherche. Pour pallier ce problème, nous nous appuyons sur le principe des sous-objectifs, autrement dit des différents thèmes (souvent distincts) que doit aborder l'utilisateur dans sa recherche complexe. Afin de définir des sous-objectifs, nous sommes partis de l'énoncé de la tâche proposée et nous avons effectué un découpage en propositions "thématiques". Nous définissons une proposition thématique comme une proposition avec un contenu sémantique distinct et donc avec un contenu lexical spécifique. À partir de cette notion, nous avons analysé le comportement des utilisateurs dans le déroulement de la session de recherche, notamment en étudiant les variations sur : le nombre total de propositions formulées dans la session, leur ordre de traitement ou encore le nombre de propositions utilisées dans une même requête.

Nous proposons donc dans cette étude, une approche exploratoire associant variables génériques et analyse des propositions thématiques de recherche dans l'objectif de définir une typologie comportementale. Cette typologie pourra servir à la personnification des moteurs de recherche. De plus, sur un plan didactique, si nous parvenons à définir des comportements plus efficaces, il serait pertinent de les enseigner afin de rechercher efficacement sur le Web.

Dans la suite du document, nous présenterons les limites des typologies du comportement des utilisateurs existantes dans la RI dans la section 2. Nous reviendrons par la suite sur une présentation détaillée de nos données et de la méthodologie mise en place (section 3). Nous continuerons en proposant une première typologie comportementale en section 4. Pour finir, nous discuterons des avantages et des limites de notre proposition (session 5).

## 2. État de l'art : les typologies comportementales en RI

À notre connaissance, il n'existe pas de typologie du comportement des utilisateurs qui se base sur des caractéristiques observables à l'échelle de la session. Actuellement dans la littérature, nous retrouvons des typologies très générales, i.e. qui ne tiennent pas compte des spécificités de la RI, ou à l'inverse très fines concentrées sur des phénomènes locaux.

La RI a été apparentée à la résolution de problèmes en tant qu'activité qui implique des "actions successives visant à transformer la situation de son état initial vers le but" [2, p.60]. La psychologie s'est donc intéressée à la RI et a dégagé différents modèles en s'appuyant notamment sur certains heuristiques de résolution de problèmes. Par exemple le modèle *General Problem Solver* de Newell et Simon (1972) a été appliqué à la RI et repose sur 4 étapes : la représentation du problème, la sélection d'un ou des opérateurs, l'utilisation de l'opérateur choisi et l'évaluation du problème, se basant sur l'heuristique de l'analyse fin-moyen [3, 4]. Des études se sont également intéressées à l'utilisateur directement en étudiant des facteurs pouvant influencer son comportement comme l'âge ou encore le niveau de connaissance du domaine de la tâche [5, 6, 7]. Ainsi, nous savons que des utilisateurs avec des connaissances avancées sur le sujet de recherche reformulent plus facilement tous les termes de la requête, contrairement à des personnes avec un niveau moins élevé qui ajoutent ou suppriment des mots de leurs requêtes [7], ou encore que des personnes âgées formulent des premières requêtes moins élaborées que des jeunes en se calquant davantage sur l'énoncé pour formuler des mots-clés [5]. D'autres études apportent un regard sur la tâche expliquant que selon la complexité un utilisateur peut reformuler un grand nombre de fois ou abandonner plus souvent [8, 7].

Cependant ces modèles restent très généraux et reposent sur des variables génériques comme le nombre de requêtes, le nombre de mots-clés, les clics, etc. À partir de ces variables, nous connaissons des caractéristiques moyennes permettant une description simple de la session de recherche, sans prendre en considération l'enchaînement des actions constituant cette session et ne donnant pas une représentation globale de celle-ci. De plus, comme nous l'avons vu, il est très fréquent qu'en psychologie les études reposent sur l'influence de certains facteurs afin de justifier des comportements. Par conséquent, ces typologies sont trop générales sans permettre l'analyse fine de la session de recherche, mais à la fois trop spécifiques car très dépendantes du type d'utilisateur ou de la tâche.

Par opposition à ces typologies générales, nous retrouvons des typologies sur des actions précises telles que la requête. En effet, certaines études ont cherché à qualifier les requêtes comme dans la taxonomie des requêtes de Broder en fonction du besoin : informationnelle, navigationnelle ou transactionnelle [9]. D'autres études se sont intéressées à la reformulation, c'est-à-dire le passage d'une requête n à une requête n+1 répondant au même besoin [10]. Nous avons vu émerger différentes typologies notamment sémantiques comme celle de Rieh et Xie (2006) où nous trouvons la description de 4 types de reformulation : spécification, généralisation, mouvement parallèle et reformulation [11]. Nous retrouvons également des typologies formelles expliquant les transformations qu'entraînent les reformulations comme l'ajout ou la suppression d'un mot, un changement morphologique (*stemming*), la différence de casse ou d'espace [12, 13, 14]. Contrairement aux premières études, ces analyses reposent sur des actions locales pendant une session de recherche. Nous avons une étude de la variation à un moment localisé de la session de recherche. De ce fait, nous ne pouvons pas déterminer un comportement global à l'échelle de la session de recherche.

Dans le cadre de notre étude, nous essayons donc de trouver une alternative à ces typologies trop générales ou trop fines, en proposant une typologie au niveau de la session de recherche. Pour ce faire, nous prenons en considération les variables génériques utilisées fréquemment tout en apportant de nouvelles variables exploratoires sur le traitement des propositions thématiques.

## 3. Méthodologie

Dans cette partie, nous présenterons notre jeu de données ainsi que la méthodologie utilisée pour construire une typologie des comportements utilisateurs à partir d'une annotation manuelle de l'espace de recherche.

### 3.1. Présentation des données

Dans cette étude, nous travaillons à partir d'une tâche de recherche complexe provenant du jeu de données CoST [15]. Dans ce jeu de données, nous trouvons les sessions de recherche de 70 participants qui devaient répondre à 15 tâches de complexité différente dans trois domaines de recherche (l'informatique, la médecine et la psychologie cognitive), sur le moteur de recherche Google. Nous avons sélectionné la tâche de prise de décision d'informatique car nous pouvons définir des propositions thématiques à partir de l'énoncé, pouvant correspondre à des sous-objectifs de recherche. Il est bien entendu possible qu'une requête aborde plusieurs propositions à la fois. Nous proposons donc le découpage suivant réalisé *a posteriori* de la création de la tâche, en nous basant sur la décomposition syntaxique de la tâche :

*Énoncé : Dans le cadre de votre stage de master, vous avez l'objectif de (**P1 : développer un programme détecteur de plagiat**). Vous souhaitez mettre en place une (**P2 : méthodologie d'analyse de textes**) mais vous hésitez entre l'utilisation simple de la (**P3 : morphologie du texte (mots, n-grammes, phrases etc.)**) ou l'utilisation de (**P4 : ressources externes (dictionnaires, thesaurus, Word embeddings)**). Après avoir relevé (**P5 : les avantages et inconvénients de chaque type d'analyse, sélectionnez la méthode qui vous parait la meilleure en justifiant vos choix**).*

Ces propositions sont donc le reflet des grandes lignes de l'énoncé permettant la réalisation de la tâche par l'utilisateur. Après observation des données, nous avons la confirmation que ces différents éléments sont présents dans les sessions de recherche à une échelle plus ou moins importante et variables d'un utilisateur à l'autre. Nous avons pu observer à partir des données que les utilisateurs ne traitaient pas ces propositions de la même façon. Certains utilisateurs ont commencé leur recherche par une requête sur les "programmes de détection de plagiat" là où d'autres se sont intéressés à la "morphologie du texte". Nous avons observé également que des participants faisaient des "va-et-vient" entre des propositions, là où d'autres opéraient une réalisation plus linéaire. De plus, nous avons constaté que certains utilisateurs formulaient des requêtes faisant référence à d'autres éléments thématiques que ceux présents dans l'énoncé.

### 3.2. Annotation manuelle des propositions thématiques

Afin d'étudier les variations comportementales en fonction de ces propositions, nous avons effectué une annotation manuelle des sessions de recherche. Pour ce faire, à partir des 5 propositions dégagées de l'énoncé, nous avons attribué un 1 à la proposition si elle est présente dans la requête. En d'autres termes, si la requête fait directement mention d'une des propositions entièrement ou partiellement, nous considérons que l'utilisateur traite cette partie de l'espace thématique. Nous avons également pris en compte une proposition "Autre" en précisant quel terme de la requête ne faisait pas référence aux 5 propositions précédentes.

**Table 1**
Annotation manuelle des propositions thématiques d'un extrait de session du jeu CoST

|   | **Requêtes** | **P1** | **P2** | **P3** | **P4** | **P5** | **PAutre** |
|---|---|---|---|---|---|---|---|
| 1 | programme plagiat AND morphologie AND "word embeddings" | **1** | 0 | **1** | **1** | 0 | 0 |
| 2 | avantages incovénients programme plagiat | **1** | 0 | 0 | 0 | **1** | 0 |
| 3 | fonctinonement programme plagiat | **1** | 0 | 0 | 0 | 0 | 0 |
| 4 | scholar | 0 | 0 | 0 | 0 | 0 | **1** |
| 5 | plagiarism detection program | **1** | 0 | 0 | 0 | 0 | 0 |

Pour cette annotation, nous avons annoté l'ensemble des sessions répondant à cette tâche,

soit 70 sessions regroupant 513 requêtes. Nous avons tout d'abord mesuré la faisabilité et la fiabilité de cette catégorisation en réalisant une double annotation pour 15 sessions de recherche. Pour les 5 propositions de l'énoncé, nous avons mesuré un Kappa de Cohen compris entre 0,80 et 1. Les rares cas de désaccord correspondent à la présence d'une proposition dans une requête en s'appuyant sur des mots proches pouvant correspondre à une proposition sans qu'ils soient clairement formulés dans l'énoncé (e.g. le nom "logiciel" peut être considéré comme faisant référence à la P1 avec un rapprochement possible avec le mot "programme"). Cependant, ces mesures indiquent des données fiables sur les 70 sessions annotées utilisées par la suite.

### 3.3. Observables : unités comportementales

Pour notre approche inductive, nous avons sélectionné un ensemble d'observables. Nous proposons ici une description de ces observables qui nous servent dans la constitution de nos profils en nous appuyant sur le tableau 1 pour illustrer nos explications. Nous retrouvons dans le tableau 2 une description quantitative de ces observables en tant que variables comportementales.

Pour commencer, nous nous intéressons à la **couverture des propositions**, soit le nombre de propositions différentes utilisées par un utilisateur au cours de la session de recherche. En effet, certains utilisateurs explorent l'ensemble de l'espace de recherche proposé par l'énoncé, là où d'autres se focalisent sur certains aspects. De plus, certains utilisateurs vont même plus loin en proposant des requêtes faisant référence à d'autres éléments que ceux présentés dans les propositions. Dans le cadre de notre exemple, l'utilisateur mentionne au total 5 propositions différentes au cours de la session.

Un deuxième observable défini est le **nombre moyen de propositions par requête**. Nous nous intéressons à la façon dont l'utilisateur organise ses propositions au sein d'une requête, c'est-à-dire le nombre de propositions simultanées dans une requête. Au cours de leur recherche, des utilisateurs peuvent formuler une requête faisant référence qu'à une seule proposition (e.g. dans la requête 3 avec la présence uniquement de P1). En revanche, il est possible de faire appel à plusieurs propositions comme dans la requête 1 (P1, P3 et P4). L'utilisateur formule donc en moyenne 1,6 proposition par requête sur cette session.

Nous nous intéressons également à la **persistance des propositions**. Nous définissons la persistance comme le nombre de requêtes consécutives où nous retrouvons une même proposition sous différentes formes. Il est parfois possible qu'un utilisateur formule une session entière en faisant la mention tout le long d'une même proposition. Dans notre exemple, nous observons que P1 est répétée en requête 1, 2 et 3. Nous avons donc ici une phase de persistance pour cette proposition. Dans le cadre de cette étude, nous définissons seulement la présence ou non de cette action sans prendre en considération la proposition concernée ou le nombre de fois où l'action est répétée. Dans cet exemple, la session a un score de 1 pour montrer la présence d'au moins une phase de persistance.

Pour finir, nous nous intéressons aux **intermittences des propositions**. Il est possible qu'un utilisateur fasse une rupture dans l'utilisation d'une proposition. L'utilisateur peut ne plus reformuler une requête faisant mention de cette proposition, mais il peut également la mentionner à nouveau. Dans ce cas-là, nous parlons d'intermittence. Par exemple, dans la session présentée, l'utilisateur fait une intermittence avec P1 puisqu'il revient sur celle-ci en requête 5 après l'avoir mentionnée une première fois dans les requêtes 1, 2 ou 3. Comme pour l'observable précédent, la session a un score de 1 puisqu'il existe au moins une action d'intermittence au cours de la session, sans distinction en fonction des propositions.

### 3.4. Identification d'une typologie

Nous commençons donc par décrire les variables génériques utilisées : le nombre de requêtes (i.e. la taille de la session) (dans le tableau 2 et la figure 1 - *NbReq*), la longueur moyenne des requêtes en nombre de mots (*LongReq*), le nombre de clics correspondant à l'ouverture de page Web (*NbClics*) et la durée approximative de la session[1] (*Durée*).

À ceux-là, nous ajoutons les variables définies à partir de nos observables : le nombre de proposition par session (*NbPSession*), le nombre moyen de proposition par requête (*PmoyReq*), la présence d'intermittence (*IntermittenceP*) et la présence de persistance (*PersistanceP*).

Les statistiques descriptives de ces variables sont présentées dans le tableau 2.

**Table 2**
Description quantitative des variables comportementales

|        | NbReq | LongReq | NbPSession | PmoyReq | IntermittenceP | PersistanceP | NbClics | Durée   |
|--------|-------|---------|------------|---------|----------------|--------------|---------|---------|
| Min    | 1,00  | 1,50    | 1,00       | 1,00    | 0,00           | 0,00         | 0,00    | 37,69   |
| Max    | 26,00 | 9,00    | 6,00       | 3,50    | 1,00           | 1,00         | 24,00   | 3402,04 |
| Moy    | 7,33  | 4,12    | 3,79       | 1,89    | 0,54           | 0,90         | 5,76    | 670,02  |
| E-type | 5,23  | 1,42    | 1,23       | 0,58    | 0,50           | 0,30         | 4,90    | 657,13  |
| Med    | 6,00  | 3,97    | 4,00       | 1,83    | 1,00           | 1,00         | 4,00    | 472,15  |

Nous notons une grande dispersion pour certaines variables notamment le nombre de requêtes formulées, les clics ou encore la durée de la session. Si nous nous intéressons aux variables en lien avec les propositions thématiques, nous observons que certains individus s'intéressent à une seule proposition à l'échelle de la session là où d'autres recherchent autour des 5 propositions de l'énoncé et plus encore. En moyenne, les utilisateurs organisent leur session autour de 4 propositions environ. Pour le nombre moyen de propositions par requête, nous notons qu'en moyenne les individus formulent des requêtes avec environ 2 propositions. Pour terminer, nous relevons déjà une variation précise au niveau de l'intermittence et de la persistance des propositions. Nous observons qu'environ 50% des participants réalisent au moins une intermittence d'une proposition et que 90% opèrent une phase de persistance.

Afin d'analyser les relations entre les variables pour proposer une analyse descriptive et exploratoire, nous avons décidé de réaliser une analyse multidimensionnelle en choisissant de représenter nos variables avec une Analyse en Composantes Principales (ACP) après centrage et réduction des variables, illustrée dans la figure 1 [16]. Le premier axe représente 43,7% de la variance totale. Il est fortement lié aux variables suivantes : le nombre de requêtes, le nombre de clics, de l'intermittence des propositions, du temps et du nombre de propositions dans une session. Le deuxième axe, représentant 22,4% de la variance, est lié au nombre de propositions en moyenne par requête et la longueur des requêtes. La persistance est liée au premier axe, mais elle a un poids plus important sur la troisième dimension, non représentée ici.

---

[1]Nous avons ici l'indice de temps de la dernière action effectuée par l'utilisateur. Il est donc possible d'envisager un temps de lecture supplémentaire pour obtenir le temps total de la session de recherche.

Pour mettre au jour nos profils, nous avons utilisé un clustering hiérarchique ascendant avec le critère d'agglomération de *Ward* sur une distance euclidienne. Nous proposons de distinguer 5 groupes dans cette étude à partir de l'ACP.

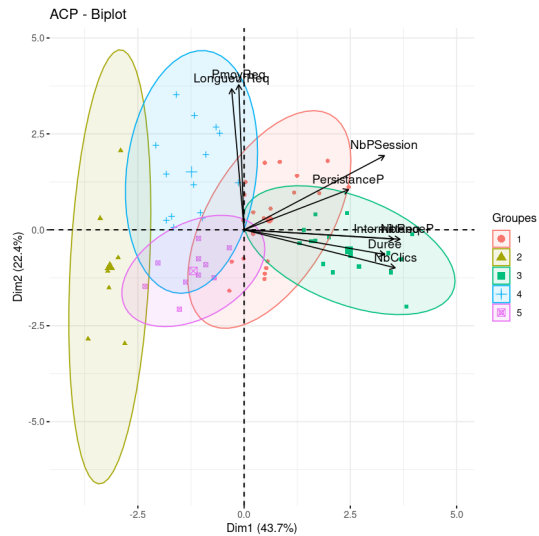

**Figure 1:** Plan des deux premières composantes principales, projection des 5 classes avec ellipses de confiance à 95%

## 4. Résultats

Malgré le recouvrement des ellipses de confiance, nous arrivons à une interprétation de 5 classes. En étudiant nos groupes, nous distinguons les individus regroupés à gauche de l'axe (les groupes 2, 4 et 5) et les individus à droite de l'axe vertical de l'ACP (les groupes 1 et 3). L'une des principales différences de cette distinction est la présence d'intermittence. À gauche, nous retrouvons des individus qui ne réalisent jamais d'intermittence, i.e. des individus qui traitent une proposition sans jamais revenir dessus pendant la session. Les individus à droite effectuent au minimum une phase d'intermittence, ils ont donc interrompu le traitement d'une proposition pour y revenir dans la suite de la session.

Parmi les individus de gauche, nous distinguons trois groupes (2, 4 et 5). À l'extrémité de notre ACP, nous trouvons le groupe 2 qui réalise des sessions très courtes composées d'une à deux requêtes ne permettant ni d'intermittence, ni de phase de persistance. Nous avons le groupe qui fournit ici le "moins d'effort". Nous différencions le groupe 4 du groupe 5 principalement sur la longueur des requêtes et le nombre de propositions moyen utilisées dans une requête et à l'échelle de la session. En effet, le groupe 4, situé au-dessus de l'axe horizontal regroupe des individus qui formulent de longues requêtes en nombre de mots. Cette longueur est justifiée sans doute par le nombre de propositions qu'ils formulent dans une même requête. Leurs sessions ne sont pas forcément longues en nombre de requêtes, ni même en durée, cependant ils recherchent autour d'un nombre assez élevé de propositions. À l'inverse, le groupe numéro 5 présente des sessions plus longues en temps et en requêtes sans pour autant rechercher beaucoup de propositions différentes avec des requêtes plus courtes également. Dans ces deux groupes, nous ne retrouvons pas de phase d'intermittence, cependant ces utilisateurs opèrent tous au minimum une phase de persistance, justifiant ainsi une distinction majeure avec le groupe 2. Il reste important de préciser que ces deux groupes réunissent des individus qui réalisent des sessions plus courtes en termes de requêtes et de durée par rapport à la moyenne des individus.

Concernant le groupe 1 et 3, à droite de l'axe vertical, nous les différencions principalement par les variables génériques. Le nombre de propositions utilisées au niveau de la session ou de la requête reste très proche pour ces deux groupes même si nous retrouvons un peu plus de propositions utilisées au niveau de la session pour le groupe 3 et au niveau de la requête pour le groupe 1. Le groupe 3 se distingue par le nombre de requêtes formulées, le temps et les clics effectués. Nous avons un groupe qui montre un effort très important sur le nombre de requêtes formulées ou le temps passé à réaliser sa session de recherche. Malgré des sessions assez longues en nombre de requête pour les individus du groupe 1, ces derniers passent beaucoup moins de temps à réaliser leur session de recherche et ne visitent que peu de pages Web. Nous avons donc un groupe 3 avec des sessions très longues marquées par la durée, le nombre de requêtes et les clics. En opposition à un groupe 1 qui a des sessions longues en nombre de requêtes uniquement et sans montrer une activité importante dans la visite de pages Web.

Pour conclure, dans l'ensemble certaines propositions semblent être décisives pour la constitution de groupes. Nous avons constaté, comme démontré précédemment, que l'intermittence de proposition divise dans la majorité les individus. En plus de cela, nous notons que le temps est un bon indicateur comme le nombre de requêtes ou encore le nombre de propositions traitées et formulées dans une même requête.

## 5. Discussion

Dans cette étude, nous avons dégagé une typologie de comportements d'utilisateurs d'un moteur de recherche sur le Web, à l'échelle de la session. Pour cela, nous avons défini et appliqué une méthode associant des variables génériques souvent utilisées dans la description des sessions à une approche exploratoire sur le traitement des propositions thématiques. Cette approche nous permet ainsi d'apporter une nouvelle façon de distinguer les utilisateurs autrement que par le nombre de requêtes formulées ou la durée de la session. Ainsi, l'observation du traitement des propositions thématiques permet une description plus fine et globale au niveau de la session.

Cependant, nous pouvons relever certaines limites à notre approche. Nos résultats dépendent du niveau de découpage de nos propositions. Pour cette étude, nous sommes assez génériques car nous voulons que notre approche soit applicable à des énoncés très différents. Pour évaluer notre typologie, nous proposons un nouveau jeu de données où les tâches ont été construites autour de cette notion de proposition thématique où nous confrontons deux propositions générales (e.g. les systèmes d'exploitation et les films d'animation).

Notre étude doit pouvoir être généralisable sans avoir besoin d'une annotation manuelle. L'annotation reste un travail chronophage pouvant comporter des erreurs humaines. Nous proposons donc d'utiliser différentes techniques (*embeddings* ou grands modèles de langue (LLM)) afin de retrouver automatiquement les propositions thématiques. Cependant, cette approche automatique soulève des questions sur le traitement des requêtes, données atypiques pour les modèles de langue.